\documentclass[runningheads]{llncs}
\usepackage{graphicx}
\usepackage{longtable}
\usepackage{newtxtext,newtxmath}

\begin{document}
\title{Bangla Image Caption Generation through CNN-Transformer based Encoder-Decoder Network}  

%

\author{Md Aminul Haque Palash\inst{1}\orcidID{0000-0002-4794-8531} \and
MD Abdullah Al Nasim\inst{1}\orcidID{0000-0001-5594-607X} \and
Sourav Saha\inst{1}\orcidID{0000-0003-0483-0581} \and
Faria Afrin\inst{1}\orcidID{0000-0001-5603-8647} \and
Raisa Mallik\inst{1}\orcidID{0000-0002-0043-8649} \and
Sathishkumar Samiappan\inst{2}\orcidID{0000-0002-8443-883X}}
%
%

\institute{Pioneer Alpha, \email{info@pioneeralpha.com} \and
Mississippi State University, USA}

%
\maketitle              
\begin{abstract}

Automatic Image Captioning is the never-ending effort of creating syntactically and validating the accuracy of textual descriptions of an image in natural language with context. The encoder-decoder structure used throughout existing Bengali Image Captioning (BIC) research utilized abstract image feature vectors as the encoder's input. We propose a novel transformer-based architecture with an attention mechanism with a pre-trained ResNet-101 model image encoder for feature extraction from images.  Experiments demonstrate that the language decoder in our technique captures fine-grained information in the caption and, then paired with image features, produces accurate and diverse captions on the BanglaLekhaImageCaptions dataset. Our approach outperforms all existing Bengali Image Captioning work and sets a new benchmark by scoring 0.694 on BLEU-1, 0.630 on BLEU-2, 0.582 on BLEU-3, and 0.337 on METEOR. 

\keywords{Bengali Image  Captioning  \and ResNet-101  \and Transformer.}
\end{abstract}
\section{Introduction}

Depiction aims to describe an image's substance in text. Plenty of the captioning algorithms now follow an architecture of an encoder-decoder where a decoder infrastructure may anticipate words using the function retrieved via a method of attention from the encoder network. Image subtitling in studies has been focused on a translation technique, including a visual encoder as well as a language decoder \cite{liu2021cptr}. The method of producing picture titles may be used for automating driving automobiles, implementing face recognition systems, describing persons with visual impairment, improving the quality of the photo query, etc. At the junction of computer vision and natural language interpretation, the challenge of producing natural language descriptions of the data of an image resides.
In some domains, for example, image search, picture captioning has already had substantial implications. It may have a positive impact in many fields, including disability software, video surveillance, and security, the interaction between humans and computers.

Image captioning seems to be the function in the natural language to describe the visual data. An algorithm is required to comprehend and represent the links underlying visual and textual information, and to create a series of the text output. The most important studies are based on the encoder-decoding framework \cite{liu2021cptr} which is quite close to the machine transcription sequence-to-sequence paradigm. The architecture of merge or mixing, which is mostly utilized for linguistic encoding. This same coded language information and visual characterize complete transformer connectivity in the Caption Transformer (CPTR) \cite{liu2021cptr} replaces a CNN with a fully convolution-free transformer encoder throughout the encoder portion. We directly sequence raw pictures as an input when compared to the traditional subtitled models that take the feature derived by CNN or feature detector as the source. Influenced by the studies above, we explore a new sequence-to-sequence approach for the resolution of the picture subtitling issue. Statistics are then fused into one multi-modal layer which predicts the term of the title by term \cite{anderson2018bottom}.


Open-domain submission seems to be a very difficult process, for it demands a thorough knowledge of the global as well as local image domains. A new operating system to evaluate image subscribing systems is a recent release of the MSCOCO project. Deep sequence model methodologies have produced amazing results. It was demonstrated recently that exposure differentiation and non-differentiable task metric questions may be solved by the incorporation of reinforcement learning approaches. 
Designers propose a novel, basic transformer network design, based only on principles of attention. At the WMT 2014, BLEU  will be delivered to the model \cite{khan2021improved}. Elements and other prominent areas of the image are a lot better natural foundation of attention to create more human-like descriptions and inquiry responses. Researchers offer a joint visual recognition mechanism from the bottom up to the top down. The findings on the MSCOCO \cite{khan2021improved} production servers provide a new cutting edge for the captioning challenge.
To summarize, the paper will contribute to these factors proceeding ahead:
\begin{itemize}
\item  We propose a novel architecture of transformer base which will be implemented for image captioning, with attention mechanism. It is tailored to the complex structure of the whole image regions \cite{khan2021improved}.

\item The model architecture will be designed using the ResNet-101 \cite{ghosal2019brain} for feature extraction from images.

\item We will perform the image depiction by using the train set and validation set of the BanglaLekhaImageCaptions dataset \cite{BanglaLekhaImageCaptions}. 


\item On the BanglaLekhaImageCaptions dataset, we evaluate our approach both qualitatively and quantitatively. 

\end{itemize}







\section{Related Works}

A rudimentary method to creating the image title may be to anticipate and connect words from image areas. Researchers built one of the earliest systems of picture subtitling in 1999. The underlying job of picture translation was subsequently retranslated in 2002. However, for various reasons, it has been shown out that the approach fails to correctly map an image to a word that entirely misses all kinds of interactions between the elements.
Earlier subtitles based on basic templates generated, filled out by an extracted features \cite{herdade2019image} or predictor attributes. With its introduction, most subtitling systems used RNNs as speech models for encoding visual input on one or more levels of a CNN \cite{herdade2019image}. New careful paradigms have been suggested, and state-of-the-art outcomes for machine interpretation and language comprehension work have been accomplished.
Self-attention would be a technique that relates several places in a specific setting to calculate a sequence presentation. It is the backbone for the Extended Neural models GPU, ByteNet, and ConvS2S \cite{khan2021improved}, all used as fundamental building blocks for deep convolutional neural network models. Throughout the Transformer, all of that is dropped to a constant number of stages. At the same time, the average attention-weighted places are lowered at the cost of such effective resolution describe the steps.
\\
In the field of image subscription, a search-based method was prominent. The system demonstrates that the spatial link \cite{wu2005novel} between items enhances the quality and the positioning of the produced annotation. One proposed approach is to integrate the pictures and text in the same place in the vector to facilitate searches. Similar efforts include using a recurring visual representation of a photograph and using deep reinforcement education \cite{wu2005novel} for picture submission. Image extraction is the second collection of data in the BanglaLekha series \cite{naim2022bangla} focused on pictures of solitary characteristics in Bangla. 
Specific direct techniques were also suggested, including the query extension method Yagcioglu et al. to cope with picture description issues \cite{bernardi2016automatic}. These techniques are creative, and each has its features, but they have the same problem of not intuitively observing the items or activities in the image. In this study, we examine the process of developing techniques of picture description in past years and describe the essential framework and improvements. The task of the decoder is to need future words with the absolute most excellent chance from all vocabulary words within the previously created title and to produce endings of the term ken. The first is the injection architecture, where the RNN is employed as a generator for the caption conditioning of the picture characteristics. Based on these fusion model accomplishments and the NLP CNN \cite{ghosal2019brain} success story, we propose an encoder-decoder-based model for Bengali Image Titling based on the fusion architecture \cite{wang2020overview}.
Contrary to the CNN used in our study, a pooling layer is used to capture essential and relevant information. We also compared our CNN with qualitative and quantitative data mixture model, and thus the CNN-LSTM  \cite{khan2021improved} data based mixture model was proposed. The CNN-based language encoder is used to improve the proposed model's overall performance.
Introducing compassionate models in which the recurring connection for self-attention is discontinued gives unparalleled set-up and sequence modeling possibilities. Multi-modal structure of the sub-titling of images for specific architectures, differing from the need for a single modality. This same transformer paradigm underlying machine translation inspires our design and contains two crucial innovations in all prior image description methods. M2 Transformer with Meshed-Memory features \cite{cornia2020meshed} consists of an encoder that includes a self-attention with a forward feeding layer stack and then a decoder that utilizes autonomous attention to the signs and awareness to the output of the final encoder phase. Tested on the COCO reference point, the "Karpathy" testing data \cite{cornia2020meshed}, the model attained a new method of state-of-the-art in both individual and set configurations. Significantly, it exceeds current online test server suggestions and ranks first between the methods provided.

\begin{figure}
\includegraphics[width=\textwidth,   height=120mm]{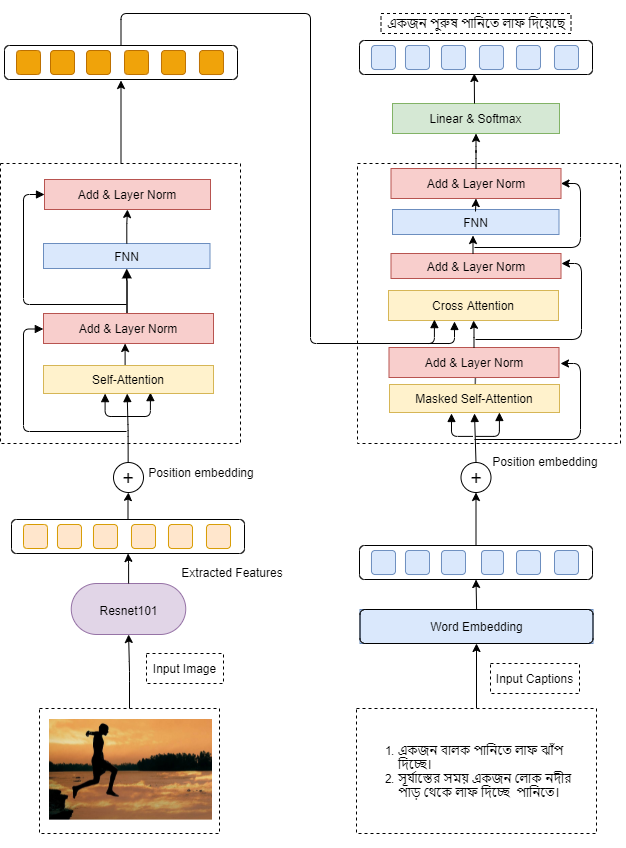}
\caption{The overview of the framework of our proposed model} \label{methodology}
\end{figure}

\section{Methodology}

We employed an Encoder-Decoder approach where the image extraction process was carried out in this research utilizing the ResNet-101 model. 
In earlier models, ResNets executed other functions, including object detection \cite{naim2022bangla} and semantical picture segmentation \cite{bernardi2016automatic}. 
In computer vision, experts have progressively substituted the conventional extractors for ResNet-101. The underlying mechanism for the efficiency of ResNets remains to be defined, though. ResNets are initially used to remove great features from the input data. 
Last but not least, the state sequence is employed to recognize the picture elements. The ResNet encoder comprises Ne stacked equivalent layers, each consisting of the sublayer and the positional forward-feed substitute. Then Positional encoding is being used to retain location coordinates as a shared component. The basic Transformer encoding methodology, \cite{vaswani2017attention} employs sinusoidal oscillations to fill out the weighted network model of positioning encoding. 
An attention layer ties all places to a predetermined sequential number of stages, whereas a recurring layer involves sequential operations. Autoattention might be limited to investigate the input data focused on the position of the respective component.
The proposed image captioning algorithm is illustrated in Fig.~\ref{methodology} as an overview. It is comprised of two parts: an encoder and a decoder.


\subsection{Encoder}

\begin{figure}
\centering
\includegraphics[width=100pt, height=140pt]{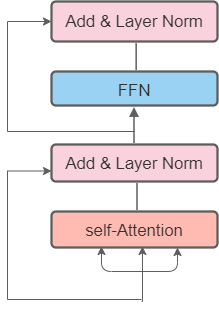}
\caption{Encoder architecture} \label{encoder}
\end{figure}

All input sequences must be translated into a consistent abstract representation that comprises all the learned information for this type of sequence by the Encoder layer. The Encoder consists of two sub-modules: multi-headed attention and another is a completely linked network. Layer normalization creates multiple residual connections between the two sublayers (see Fig.~\ref{encoder}).

\subsubsection{Feature Extraction:}
As an image feature extractor, we utilized the pre-trained ResNet-101 \cite{chen2021pre}. The ImageNet dataset \cite{rennie2017self} was used to train it. Pattern recognition has always been a strong suit for neural networks with several layers. There are overfitting problems and they are difficult to optimize. Residual CNNs are built with fast connections between layers in mind. Identity mapping is performed via these relationships. ResNets are simple to tune, and their implementation improves as the network depth grows. We discarded the ResNet-101's final output layer, which contains the picture classification output, and only used the encoded image characteristics created by the hidden layers.

\subsubsection{Positional Encoding:}

We must provide data about the relative or absolute positions of the tokens in the series for the model to use the sequence order because ours lacks recurrence and convolution. To accomplish so, we add "positional threescore encodings" to the bottoms of the input embeddings of the encoder and decoder stacks.

There are a variety of learned and fixed positional encodings from which to choose.
We utilize sine and cosine functions of various frequencies in this work:
Let t stand for the desired location in an input sentence, be the encoding that corresponds to it, and d stand for the encoding dimension, 
$ \overrightarrow{P_{t}} \; \epsilon \; \mathbb{R}^{d} $
be its corresponding encoding, and d be the encoding dimension 
 (where $ d \equiv 20 $) then $ f : \mathbb{N\rightarrow} \mathbb{R}^{d} $ will be the function that produces the output vector $ \overrightarrow{P_{t}} $ and it defined as follows:
 
 
\begin{equation}
\overrightarrow{p_{t}}^{(i)}=f(t)^{(i)}  \label{equ pt}
\end{equation}
Here, the limit is between $ \sin(w_{k}.t), \mathbf{if} \;  \mathit{i}=2k \; and \;  cos(w_{k}.t), \mathbf{if} \; \mathit{i}=2k+1 $
 \\ where
\begin{equation}
\omega _{k}=\frac{1}{10000^{2k/d}} \label{equ ommega}
\end{equation} 

The frequencies decrease along the vector dimension, as can be seen from the function description. As a result, the wavelengths follow a geometric progression from $2\pi$ to 10000$\cdot 2\pi$ on the wavelengths. The positional embedding pt may alternatively be thought of as a vector containing pairs of sines and cosines positional $ embedding \rightarrow  pt$  for each frequency (note that d is divisible by 2).


\subsubsection{Multi-Headed Attention:}
\begin{figure}
\centering
\includegraphics[width=150pt, height=200pt]{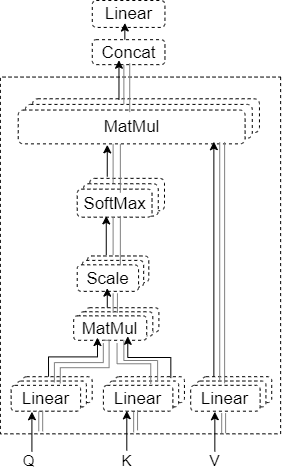}
\caption{Multi-Headed Attention mechanism} \label{Multi-Headed Attention}
\end{figure}

Multi-headed attention in the encoder uses an unique attention technique called self-attention model. Self-attention, in our case, allows models to connect each pixel in the input three-three to other pixels (see Fig.~\ref{Multi-Headed Attention}). Where the Query, Key, and Value Vectors are Q, K, and V, respectively. To achieve self-attention, we feed the input into three distinct, ultimately linked layers to produce the query, key, and value vectors.

\subsubsection{Dot Product of Query and Key:}
The questions and keys are multiplied by a dot product matrix to produce a scoring matrix after passing the query, key, and value vectors through a linear layer. The scoring matrix indicates how much importance a pixel should obtain in comparison to other pixels. As a result, each pixel in the time step will be assigned a score related to the scores of other pixels in the time step (see Fig.~\ref{dot-product}). The higher the score, the greater the concentration.


\subsubsection{Scaling Down the Attention Scores:}

\begin{figure}
\centering
\begin{minipage}{.5\textwidth}
  \centering
  \includegraphics[width=.8\linewidth,  height=80pt]{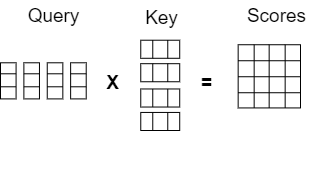}
  \caption{Dot product with query and key metrics}
  \label{dot-product}
\end{minipage}%
\begin{minipage}{.5\textwidth}
  \centering
  \includegraphics[width=.8\linewidth,height=74pt]{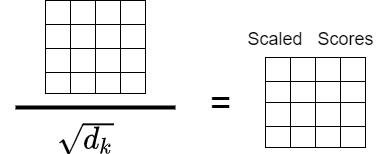}
  \caption{Scaled score metrics}
  \label{scale-down}
\end{minipage}
\end{figure}

After that, the scores are scaled-down by dividing them by the square root of the query and critical dimensions (see Fig.~\ref{scale-down}). This score allows for more steady gradients because multiplying numbers can have explosive results.After that, the scores are scaled-down by dividing them by the square root of the query and critical dimensions. This score allows for more steady gradients because multiplying numbers can have explosive results.

\subsubsection{Softmax of the Scaled Scores Multiply Softmax Output with Value vector:}
The softmax of the scaled score is then used to determine the attention weights, yielding probability values ranging from 0 to 1. Your higher scores rise while your lower scores decline when you do a softmax. It makes choosing which pixel to focus on for the model much more accessible.
\begin{equation}
Z = softmax(\frac{Q*K^{T}}{\sqrt{d_{k}}})*V   \label{equ 4}
\end{equation} 
Where Z is the self-attention matrix, Q is the query matrix, K is the fundamental matrix, V is the value matrix, and dk is the primary matrix's dimension. \cite{vaswani2017attention} improved the self-attention layer by incorporating a "Multi-Head" attention mechanism. The self-attention computation is implemented eight times with various weight matrices in Multi-Head attention.

\subsubsection{The Residual Connections, Layer Normalization, and Feed Forward Network:}
The residual connection is formed when the multi-headed attention output vector is cumulated to the original positional input embedding. The output of the residual link is normalized at the layer level before being projected for further processing using a pointwise feed-forward network. A ReLU activation separates two linear layers in the pointwise feed-forward network. The output is then normalized by adding it to the input of the pointwise feed-forward network. By allowing gradients to pass straight through the networks, the remaining links improve network training. The grid is stabilized using layer normalizations, resulting in a considerable reduction in training time. The pointwise feed-forward layer is used to project better possible attention outcomes.

\begin{equation}
FFN(x)=FC_{2}(Dropout(ReLU(FC_{1}(x))))   \label{equ 5}
\end{equation} 
A sublayer connection in each sublayer is made up of a residual link proceeded by layer normalization.
\begin{equation}
x^{out}= LayerNorm(x^{in}+Sublayer(x^{in}))   \label{equ 6}
\end{equation}
Where $x^{in}$, $x^{out}$ are the input and output of one sublayer,feed-forward and the sublayer can be the attention feed-forward-forward layer.

\subsection{Decoder}

On the decoder side, we combine the word embedding features with sinusoid positional embedding and use both the extra findings and encoder output features as input. The decoder is made up of Nd identical layers, each of which has a mask multi-head self-attention sublayer and a multi-head cross attention sublayer, as well as a positional feed-forward sublayer in that order. A linear layer with an output dimension identical to the vocabulary size uses the output feature throughout the last decoder layer to figure out the meaning of the next word. Given a ground truth sentence $ y_{1:T}^{*} $ and the prediction $ y_{t}^{*} $ of the captioning model with parameters $  \Theta $, we minimize the subsequent cross-entropy loss:

\begin{equation}
L_{XE}(\Theta)= -\sum_{t=1}^{T}log(p_{\Theta}(y_{t}^{*}|   y_{1:t-1}^{*}))   \label{equ 7}
\end{equation} 
We used the sahajBERT model from the Hugging Face library for word embedding in the decoding stage.
Like other encoding processes, we have used multi-head self-attention sublayer, multi-head cross attention sublayer, and positional feed-forward sublayer in the decoding stage to finetune our model.

\section{Experiments}


\subsection{Datasets}

This part mainly focuses on methods for analyzing open-source datasets and generated sentences in this domain, where a good dataset might help an algorithm or model perform better.

We have tested and trained our model on the BanglaLekhaImageCaptions dataset \cite{BanglaLekhaImageCaptions}.There are 9,154 images in the dataset presented are collected from the public domain. Besides, it is relevant to Bengali culture to some extent.
While this collected data is small in comparison to other datasets where the prominent image captioning datasets such as MSCOCO \cite{chen2015microsoft}, PASCAL 1K \cite{rashtchian2010collecting}, Flickr8k \cite{hodosh2013framing}, Flickr30k \cite{plummer2015flickr30k} contain a strong western cultural bias, with annotations written in English.

To train our model, we split the data set into two parts: train and validation where we have used 2 captions per image. We used 7323 x 2 images for the training. The remaining 1831 x 2 images were used in during validation.

\subsection{Data Preprocessinng}

A considerable amount of human bias exists in the BanglaLekhaImageCaptions dataset \cite{BanglaLekhaImageCaptions}. Any model's capacity to describe non-human subjects is hampered because of this bias. In certain circumstances, the captions aren't as descriptive as they should be, resulting in inaccurate training and evaluation of any model. For this reason, we have done some preprocessing to ensure that our dataset is noise-free and that it can perform well. 

Even though the Hugging Face library has a large number of pre-trained Bangla models and for preprocessing we have used sahajBERT \cite{Khalid_2021} which is a pre-trained language model that consists of 
\begin{itemize}
\item A tokenizer created specifically for Bengali and 
\item An ALBERT architecture that has been cooperatively pre-trained on a dump of Wikipedia in  Bengali and the Bengali section of OSCAR.
\end{itemize}
By fine-tuning their pre-trained models three times on two downstream tasks in Bengali. Besides, we assess the quality of the sahajBERT model and two other model benchmarks (XLM-R-large and IndicBert).



\section{Evaluation}

\subsection{Quantitative Analysis}

For Quantitative results, there are five evaluation indexes, namely, BLEU \cite{papineni2002bleu}, METEOR \cite{denkowski2014meteor}, ROUGE \cite{lin2004rouge}, CIDEr \cite{vedantam2015cider}, and SPICE \cite{anderson2016spice} to evaluate the models' predicted caption.


Among all five indicators, BLEU \cite{papineni2002bleu} is the most often used criterion for evaluation. BLEU considers translation length, word order, and word choice when determining how near a machine translation is to a human reference translation. It is used for image captioning, machine translation, etc. 

METEOR \cite{denkowski2014meteor} is also used to evaluate machine translation, in which the model's generated translation is compared to the reference translation. It's intended to address some of the issues with BLEU. 

ROUGE \cite{lin2004rouge} is an excellent metric for assessing the performance of machine translation tasks and automatic summarization. The better the performance, the higher the RUGE score.

CIDEr \cite{vedantam2015cider} was designed primarily to solve image annotation issues. 
It aims to address the case of a weak correlation between previous metrics and human judgments.

SPICE \cite{anderson2016spice} is a systematic metric for automatic evaluation comparing propositional semantic content in image captions. Rather than using the traditional n-gram metrics, SPICE is better at capturing human opinions about the model's subtitles on the dataset of raw image captions.

Here, higher metric results suggested superior performance in the BLEU and METEOR indexes.
In Table~\ref{table 1}, the scores have been highlighted with the highest accuracy in boldface where a comparative study among our model, CNN language encoder with a merged-based architecture \cite{khan2021improved}, LSTM language encoder with a mixture-based architecture \cite{tanti2017role}, and Bi-LSTM language encoder with an inject-based architecture \cite{rahman2019chittron} can be found here. 


\begin{longtable}{|l|l|l|l|l|l|}
\caption{Comparing the performance of various models quantitatively}\label{table 1} \\
\hline
\multicolumn{1}{|c|}{\textbf{Model}} & \multicolumn{1}{c|}{\textbf{BLEU-1}} & \multicolumn{1}{c|}{\textbf{BLEU-2}} & \multicolumn{1}{c|}{\textbf{BLEU-3}} & \multicolumn{1}{c|}{\textbf{BLEU-4}} & \multicolumn{1}{c|}{\textbf{METEOR}} \\ \hline
\endhead
Our proposed model & \textbf{0.694} & \textbf{0.580} & \textbf{0.505} & \textbf{2.22e-308} & \textbf{0.337} \\ \hline

CNN + ResNet-50 [merged] \cite{khan2021improved}                    & 0.651                                 & 0.426                                 & 0.278                                 & 0.175                                 & 0.297                                  \\ \hline
CNN + LSTM [mixture] \cite{tanti2017role}                                 & 0.632                                 & 0.414                                 & 0.269                                & 0.168                                 & 0.291                            \\ \hline
CNN + Bi-LSTM [inject] \cite{rahman2019chittron}                                 & 0.619                                 & 0.403                                  & 0.261                                   &  0.163                                 & 0.296                  \\ \hline
\end{longtable}

As can be seen, our proposed model achieved better results than the traditional architecture. It's also worth mentioning that the BLEU and METEOR scores are higher, indicating that our captions are of higher quality.

\subsection{Qualitative Analysis}

Fig.~\ref{all in one} depicts a qualitative example of our model's performance.
While the assessment metrics provide a numerical indication of the validity of the captions, any result can be misinterpreted.
Detecting subtle differences in the generated captions compared to the original human language description can be evaluated via qualitative analysis.
This concept will be bolstered even more by the qualitative assessment offered below.

\begin{longtable}{l}

\begin{minipage}{1.0\textwidth}
      \includegraphics[width=\textwidth, height=60mm]{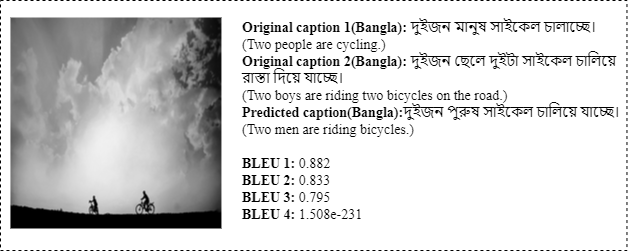}
    \end{minipage} 
\\

\begin{minipage}{1.0\textwidth}
      \includegraphics[width=\textwidth, height=60mm]{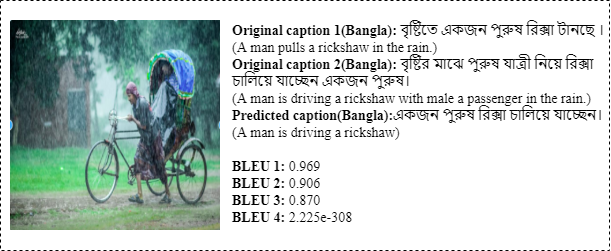}
    \end{minipage}  \\
\begin{minipage}{1.0\textwidth}
      \includegraphics[width=\textwidth, height=60mm]{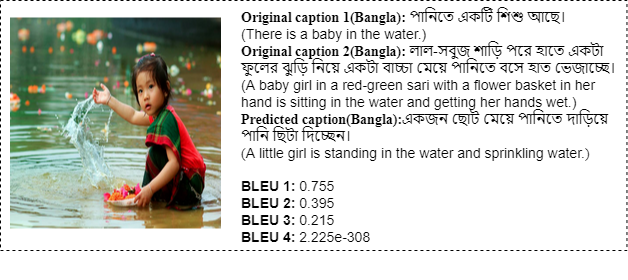}
    \end{minipage}  \\
\begin{minipage}{1.0\textwidth}
      \includegraphics[width=\textwidth, height=60mm]{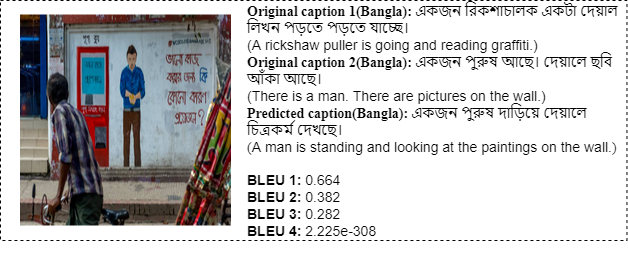}
    \end{minipage}  \\
\begin{minipage}{1.0\textwidth}
      \includegraphics[width=\textwidth, height=60mm]{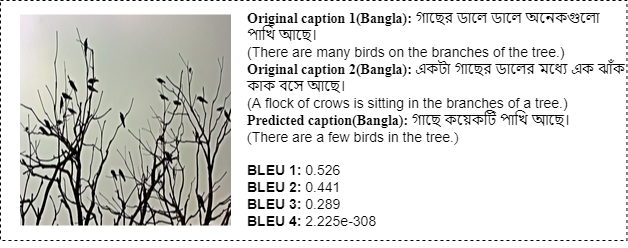}
    \end{minipage}  \\
\begin{minipage}{1.0\textwidth}
      \includegraphics[width=\textwidth, height=60mm]{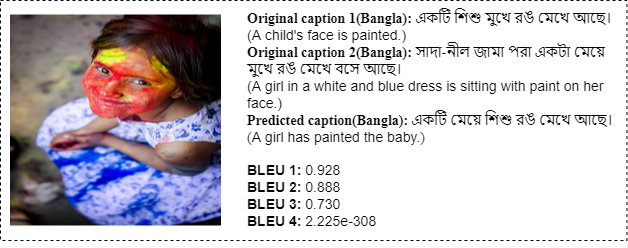}
    \end{minipage}  \\
\begin{minipage}{1.0\textwidth}
      \includegraphics[width=\textwidth, height=60mm]{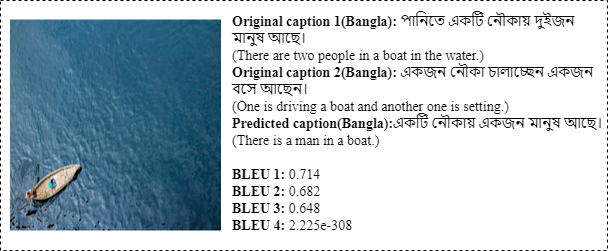}
    \end{minipage} 

\end{longtable}

\begin{figure}
\caption{Qualitative examples of Bengali captions generated by our proposed model with untouched captions and BLEU score}
\label{all in one}
\end{figure}

For non-native Bengali speakers, we provide the predicted Bengali caption and the associated English Translated caption. Our algorithm outperformed the competitive edge not only in terms of scoring but also in terms of caption quality. In comparison to other models, our model correctly identified the most appropriate caption as well as very well-detailed.



\section{Conclusion and Future Work}

This paper introduces a novel Transformer-based architecture using a Deep Neural Network for automatically developing captions in Bengali from an input image, with a comprehensive test on the BanglaLekhaImageCaptions dataset, which yielded excellent results, justifies the feasibility of our proposed approach. Besides, the BLEU and METEOR scores were used to evaluate the proposed architecture's performance and the results, which were analyzed using both quantitative and qualitative methods.
The result shows that image captioning in the Bengali language can be improved shortly.
Furthermore, we will aim to adapt the visual attention and transformer models instead of ResNet-101 for improved feature extraction and more precise captions. We anticipate that our approach will encourage others to develop more efficient transformer-based architectures to aid image captioning as well as other computer vision tasks requiring relational attention.

\bibliographystyle{splncs04}
\bibliography{bibliography}
%





\end{document}